\documentclass[sigconf]{acmart}
\PassOptionsToPackage{prologue,cmyk}{xcolor}

\AtBeginDocument{%
  \providecommand\BibTeX{{%
    Bib\TeX}}}

\copyrightyear{2023} 
\acmYear{2023} 
\setcopyright{acmlicensed}\acmConference[MM '23]{Proceedings of the 31st
ACM International Conference on Multimedia}{October 29-November 3,
2023}{Ottawa, ON, Canada}
\acmBooktitle{Proceedings of the 31st ACM International Conference on
Multimedia (MM '23), October 29-November 3, 2023, Ottawa, ON, Canada}
\acmPrice{15.00}
\acmDOI{10.1145/3581783.3611728}
\acmISBN{979-8-4007-0108-5/23/10}

\usepackage[ruled,linesnumbered]{algorithm2e}
\usepackage{multirow}
\usepackage{subfigure}
\usepackage{graphicx}
\usepackage{colortbl}
\usepackage{color}
\usepackage[cmyk]{xcolor}
\usepackage{hyperref}
\hypersetup{
            colorlinks=true,
            linkcolor=blue,
            anchorcolor=blue,
            citecolor=blue}
\definecolor{mygray}{gray}{0.85}
\begin{document}

\title{DeNoising-MOT: Towards Multiple Object Tracking with Severe Occlusions}

\author{Teng Fu}
\affiliation{%
  \institution{Shanghai Key Laboratory of IIP}
  \institution{School of Computer Science,\\Fudan University}
  \city{Shanghai}
  \country{China}}
\email{fut21@m.fudan.edu.cn}

\author{Xiaocong Wang}
\affiliation{%
  \institution{Shanghai Key Laboratory of IIP}
  \institution{School of Computer Science,\\Fudan University}
  \city{Shanghai}
  \country{China}}
\email{xcwang20@fudan.edu.cn}

\author{Haiyang Yu}
\affiliation{%
  \institution{Shanghai Key Laboratory of IIP}
  \institution{School of Computer Science, Fudan University}
  \city{Shanghai}
  \country{China}}
\email{hyyu20@fudan.edu.cn}

\author{Ke Niu}
\affiliation{%
  \institution{Shanghai Key Laboratory of IIP}
  \institution{School of Computer Science,\\Fudan University}
  \city{Shanghai}
  \country{China}}
\email{kniu22@m.fudan.edu.cn}

\author{Bin Li}
\affiliation{%
  \institution{Shanghai Key Laboratory of IIP}
  \institution{School of Computer Science,\\Fudan University}
  \city{Shanghai}
  \country{China}}
\email{libin@fudan.edu.cn}

\author{Xiangyang Xue}
\affiliation{%
  \institution{Shanghai Key Laboratory of IIP}
  \institution{School of Computer Science,\\Fudan University}
  \city{Shanghai}
  \country{China}}
\email{xyxue@fudan.edu.cn}
\authornote{Corresponding author}

\renewcommand{\shortauthors}{Teng Fu et al.}
%% No italics and no comma
%% If needed use a foot or author note to identify equal contribution

\begin{abstract}
Multiple object tracking (MOT) tends to become more challenging when severe occlusions occur. In this paper, we analyze the limitations of traditional Convolutional Neural Network-based methods and Transformer-based methods in handling occlusions and propose DNMOT, an end-to-end trainable DeNoising Transformer for MOT. To address the challenge of occlusions, we explicitly simulate the scenarios when occlusions occur. Specifically, we augment the trajectory with noises during training and make our model learn the denoising process in an encoder-decoder architecture, so that our model can exhibit strong robustness and perform well under crowded scenes.
Additionally, we propose a Cascaded Mask strategy to better coordinate the interaction between different types of queries in the decoder to prevent the mutual suppression between neighboring trajectories under crowded scenes. Notably, the proposed method requires no additional modules like matching strategy and motion state estimation in inference. 
We conduct extensive experiments on the MOT17, MOT20, and DanceTrack datasets, and the experimental results show that our method outperforms previous state-of-the-art methods by a clear margin.
\end{abstract}





\maketitle

\begin{figure}
  \centering
  \includegraphics[width=\linewidth]{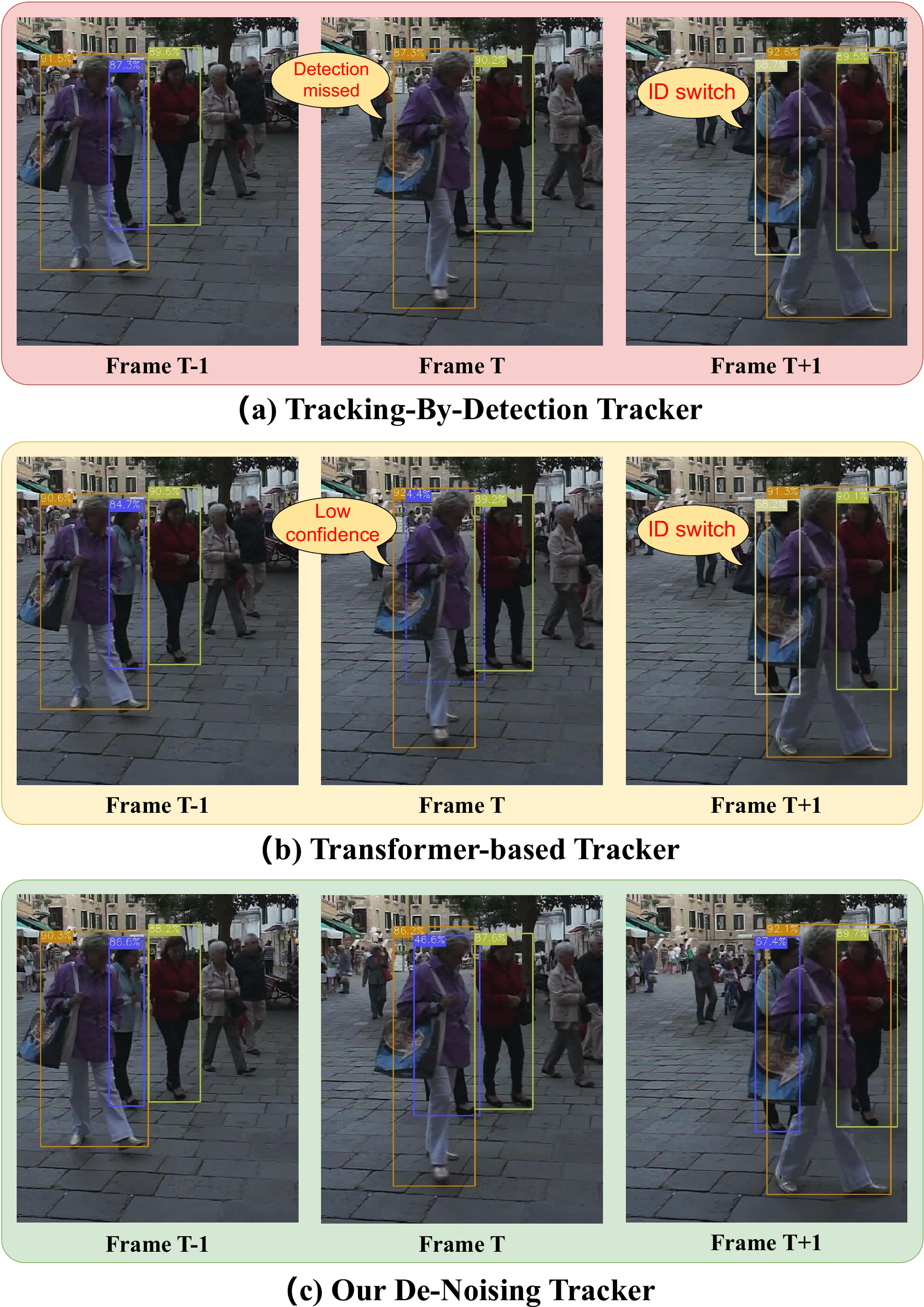}
  \caption{Visualization of a tracklet in the MOT17 dataset. Our model is robust to severe occlusions through Cascaded Mask Module and Denoising Training.}
  \label{figure1}
\end{figure}

\section{Introduction}

Multiple Object Tracking (MOT)~\cite{feichtenhofer2017detect,zhang2021fairmot,bergmann2019tracking} is a fundamental computer vision task that involves predicting the trajectory of each object in a continuous image sequence while maintaining consistent object identity~\cite{kim2015multiple,zhang2020long,bochinski2017high,berclaz2011multiple}. In recent years, MOT has found wide application in areas such as autonomous driving and video surveillance. However, due to the complexity and crowdedness of real-world scenarios~\cite{stewart2016end,alahi2016social}, severe or even complete occlusions between objects are common. As a result, preventing ID switches under severe occlusions has become a critical challenge in MOT.

Recently, the Tracking-by-Detection (TBD) paradigm~\cite{nasseri2023online,zhou2020tracking,wojke2017simple,bewley2016simple,bergmann2019tracking} has become the mainstream method for MOT due to its excellent efficiency and effectiveness. 
%These methods utilize a pre-trained detection model to detect the objects in each frame of image sequences and then match the results of detection and trajectories using motion or Re-ID information~\cite{wang2020towards}. 
The paradigm benefits from the rapid development in the field of object detection and heavily relies on the detector's performance~\cite{ge2021yolox,redmon2016you,ren2015faster,zhou2019objects}. However, when severe occlusion occurs, the object may become almost invisible in the 2D image, making it challenging to obtain the object's bounding box through the detection model. As shown in Figure \ref{figure1}(a), the middle person is occluded, and no detection result can be obtained, resulting in a failure to participate in the subsequent matching process. When the person reappears, a new trajectory is initiated.

Since the proposal of the Transformer~\cite{vaswani2017attention} model in natural language processing, the attention model~\cite{sutskever2014sequence} has rapidly emerged in the field of computer vision~\cite{dosovitskiy2020image,liu2021swin}.
% , surpassing traditional CNN-based methods in terms of performance. 
Following the success of the DETR methods~\cite{carion2020end,liu2022dab,li2022dn,sun2021sparse} in object detection, this structure has gradually been applied to multi-object tracking tasks. 
%Similar to the detection task, a detection query is used to detect new trajectories, while a separate query is reserved for each active trajectory to predict its position in the current frame. 
As illustrated in Figure \ref{figure1}(b), each existing trajectory will eventually receive a prediction result. However, due to occlusion-induced invisibility, these trajectories tend to predict lower confidence scores, leading to terrible results similar to those obtained in TBD methods.

\begin{figure}
  \centering
  \includegraphics[width=\linewidth]{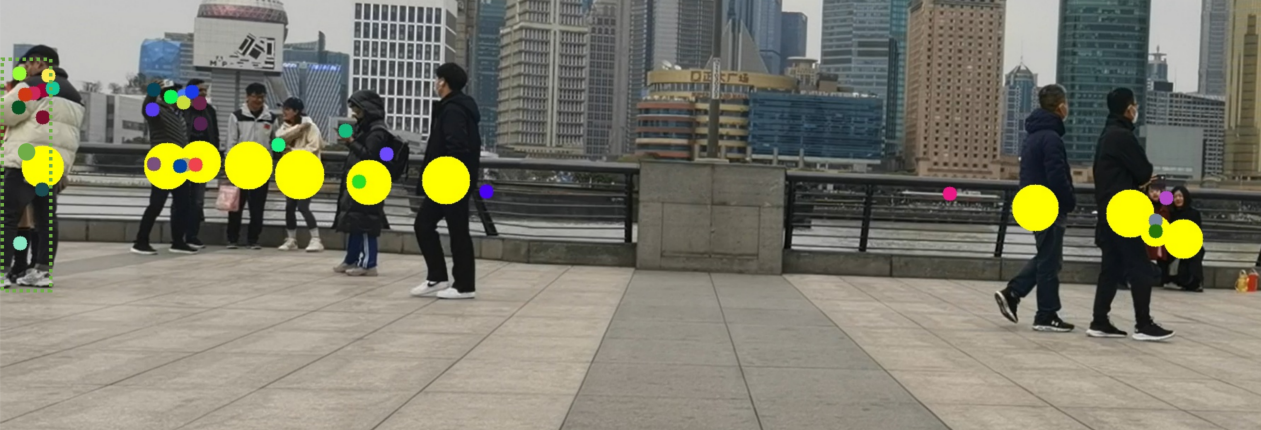}
    \caption{The reason that DETR-based models do not need NMS. We visualized all the queries whose prediction category is "person". We use yellow circles to represent the predictions that are not filtered out (the filtering threshold is 0.3), and use small circles of different colors to indicate results with low confidence.}
  \Description{Why DETR-based models do not need NMS. }
  \label{figure2}
\end{figure}

We further investigate the underlying mechanisms of the Transformer and observe that DETR~\cite{carion2020end} does not require Non-Maximum Suppression for post processing. Instead, it filters all outputs based on confidence and achieves excellent detection results. As illustrated in Figure \ref{figure2}, we can see that there is not always a one-to-one correspondence between the queries and the actual objects. Instead, multiple queries often attend to the same object. However, after the self-attention layers in the decoder, a query will suppress the neighboring queries. In the 2D-MOT task, due to the perspective relationship, objects are frequently subject to severe or even complete occlusion. This automatic suppression mechanism will also affect the queries responsible for two closely located objects. Ultimately, the confidence of the query responsible for the occluded object becomes lower, leading to the object being filtered out.

In this paper, we propose DeNoising MOT, abbreviated as DNMOT, an online object tracker based on Transformer. Our fully end-to-end trainable model needs no additional module such as Kalman Filter~\cite{kalman1960new} and Hungarian Matching~\cite{kuhn1955hungarian} in inference. To improve the model's ability to handle occlusions, we introduced Denoising Training and Cascaded Mask Module. As shown in Figure \ref{figure1}(c), our model exhibits better robustness against severe occlusions and maintains the tracking state of the object to prevent ID-Switch when it reappears.

The proposed method, based on DINO ~\cite{zhang2022dino}, first encodes the multi-scale features extracted by the CNN~\cite{simonyan2014very,he2016deep}. And then, as in the previous methods~\cite{meinhardt2022trackformer,zeng2022motr}, we use track queries and detection queries to track existing trajectories and detect new trajectories, and introduce a Query Selection Module to give more reliable location prior for detection query. Subsequently, we create another type of queries, named denoising queries, to simulate the occurrence of occlusions. Specifically, we perform three types of noise to the locations of all objects in the ground truth based on whether there are other bounding box overlaps. The denoising queries are then fed into the decoder along with other queries to improve the model's resistance to noise. Finally, to coordinate the interaction between different types of queries, we proposed a Cascaded Mask Module, which can help the queries in the decoder focus on their own trajectories and not be suppressed by neighboring trajectories. Importantly, the denoising query is not required in inference, and only two adjacent frames are needed as input in each time step, which significantly improves the time and space efficiency of our model.

The experimental results of DNMOT on MOT Challenge~\cite{milan2016mot16,dendorfer2020mot20,voigtlaender2019mots} and DanceTrack~\cite{sun2022dancetrack} demonstrate that our model achieves the state-of-the-art performance among all end-to-end Transformer-based methods. These results highlight the importance of solving the occlusion problem, not only for enhancing application performance, but also for imporving metrics such as MOTA and HOTA~\cite{bernardin2008evaluating, luiten2021hota}.
Additional ablation experiments also illustrate the effectiveness of the proposed method.

\section{Related Work}
% In this section, we briefly introduce two paradigms commonly used in multi-object tracking, Tracking-By-Detection (TBD) and Transformer-based methods, and briefly explain how existing methods deal with the occlusion problem and their shortcomings.

In this section, we will provide a brief introduction to two paradigms commonly used in multi-object tracking, namely Tracking-By-Detection (TBD) methods and Transformer-based methods. We will also briefly explain how existing methods deal with the occlusion problem and their limitations.

\subsection{Tracking-By-Detcetion Methods}

Most multi-object tracking models based on convolutional neural networks adopt the Tracking-By-Detection (TBD) paradigm, which usually consists of two stages. Firstly, a detection model is used to detect objects in the frame, and then an association stage is performed on existing trajectories and detection results using motion information~\cite{sharma2018beyond} or Re-ID information~\cite{wang2020towards}. SORT~\cite{bewley2016simple} uses the Hungarian matching algorithm and Kalman filter to achieve these two processes, and Deep SORT~\cite{wojke2017simple} adds Re-ID features in the association stage. Subsequent models focus on improving association strategies~\cite{dai2021learning,wang2021joint,andriyenko2011multi}, more robust action prediction~\cite{saleh2021probabilistic,yang2021remot}, and more efficient Re-ID feature extraction~\cite{yu2022relationtrack,shuai2021siammot,ristani2018features}. For example, ByteTrack~\cite{zhang2022bytetrack} uses more detection results and a two-stage association process; OC SORT~\cite{cao2022observation} iteratively updates the Kalman Filter through interpolation after the object reappears, while BoT-SORT~\cite{aharon2022bot} estimates the motion of the camera to correct the prediction of the Kalman filter. C-BIOU Tracker~\cite{yang2023hard} uses buffered IOUs in the association process to increase the probability of successful association. However, these methods still struggle to handle False Negative objects. 

\subsection{Transformer-Based Methods}

With the increasing popularity of the Transformer architecture in Natural Language Processing, this structure is now being widely used in Computer Vision tasks~\cite{liu2021swin,dosovitskiy2020image,yu2021benchmarking}, and achieves comparable and even state-of-the-art results in some fields~\cite{wang2023omnitracker,yu2022chinese,chu2023transmot,shi2016end}. In multi-object tracking, TransTrack~\cite{sun2020transtrack} replaces the detection model and location prediction model in the TBD paradigm with a transformer architecture, while Trackformer~\cite{meinhardt2022trackformer} uses detection query and track query to detect new trajectories and track existing trajectories respectively. MeMOT~\cite{cai2022memot} utilizes a memory buffer to enable the model to use both short-term and long-term information for better inference capabilities. These methods completely discard external modules such as the Kalman Filter used for location prediction in previous methods~\cite{chen2022tr,zhou2022global} in inference. Moreover, with the recent advancements in multimodal models, some models also incorporate information from other modalities to enhance performance~\cite{wu2023referring,zhu2023visual, dendorfer2022quo}. Our proposed DNMOT is also based on a fully end-to-end Transformer architecture and is equipped with a Denoising Training process and Cascaded Mask Module, which improves the model's robustness to occlusions and leads to better performance on evaluation metrics.

\subsection{Occlusion Handling}

Severe occlusion usually leads to problems such as missed detections and ID switches, which seriously affect the performance of the model in crowded scenes. In recent years, researchers have started to address the issue of occlusion in multi-object tracking~\cite{xu2021transcenter}. While using ReID information to match reappearing objects is a common method to handle occlusion~\cite{stadler2021improving}, it does not work well for scenarios where the object is often locally invisible~\cite{caoobject}. To address this issue, MotionTrack~\cite{qin2023motiontrack} uses the Interaction Module to model the relationship between tracks for better results in dense scenes; FineTrack~\cite{ren2023focus} uses locally unoccluded parts for fine-grained feature extraction; P3AFormer~\cite{zhao2022tracking} uses a point-wise approach to solve the occlusion problem at the pixel level; Some methods~\cite{nasseri2023online} also use calculation in the case of occlusion to determine whether the trajectory is terminated, rather than relying on inactivity time. In contrast, our DNMOT uses a novel approach to simulate the occlusion occurrence by the noising method, enabling the model to learn the denoising process when occlusion occurs. This novel approach has shown to be highly effective in addressing the challenges associated with occlusion, and its effectiveness surpasses that of previous techniques. Our experiments show that this explicit treatment, which is completely different from previous methods, leads to better robustness of the model to occlusions. 

\section{Methodology}

\begin{figure*}
    \centering
    \includegraphics[width=1.0\textwidth]{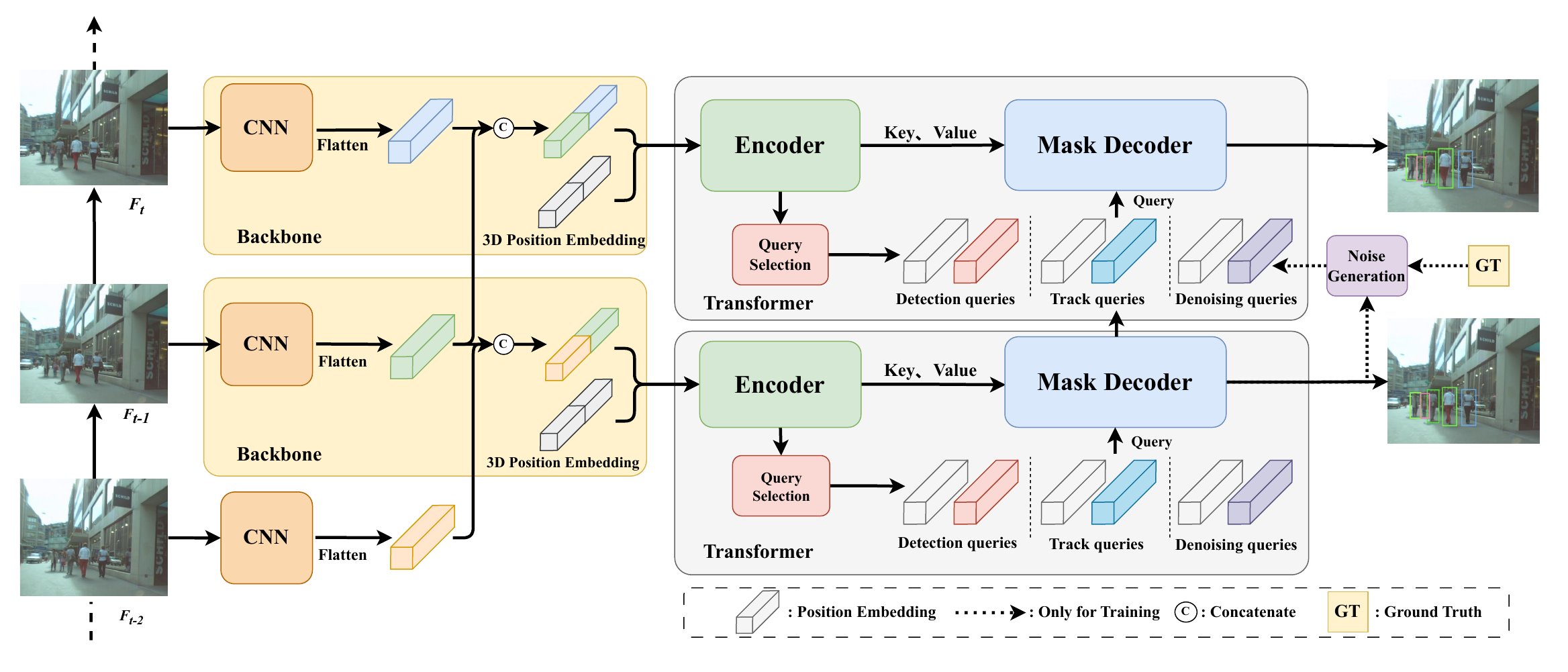}
    \caption{Overall architecture of the proposed method, consisting of a backbone for feature extraction and an Encoder-Decoder architecture. At each time step, the multi-scale features extracted from the input image will be flattened and concatenate with the features of the previous frame, and after the Encoder processing, the features will be selected to obtain the positions of interest. These positions will form detection queries with learnable embeddings, and then the track queries generated in the previous frame and the output of the Noise Generation module (Denoising queries) will be used as a Query set to pass into the Mask Decoder. Noise Generation will not be enabled during Inference.}
    \label{figure3}
\end{figure*}

Given a sequence of video frames $\boldsymbol{I}=\left\{I^0, I^1, \cdots, I^t\right\}$, the proposed DNMOT model processes each frame sequentially and generates $K$ trajectories $\boldsymbol{T}=\left\{T_0,T_1,\cdots,T_K\right\}$. 
%Each trajectory is represented by a number of five-tuples $(t,x_t,y_t,w_t,h_t)$, where $t$ represents the frame number, and $(x_t, y_t)$ represents the center point coordinates of the object in $t$-th frame, and $w_t$ and $h_t$ represent the width and height of the object in $t$-th frame, respectively.

The pipeline of our method is illustrated in Figure \ref{figure3}. Our model consists of a Backbone that extracts multi-scale features and a Transformer structure that performs tracking. The extracted multi-scale features are flattened and concatenated before being encoded using self-attention in Transformer encoder. Then the encoded features are sent to the decoder as Key and Value, and a Query set containing Detection queries, Track queries, and Denoising queries is fed to the decoder simultaneously. After the cascaded mask decoder, these queries are used to initialize new trajectories, track active trajectories, and train the model to be robust to noises, respectively. 
We will explain how to generate the three types of queries in Section \ref{sec:qs}, while our network architecture and the Cascaded Mask Module it contains will be explained in Section \ref{sec:na} and Section \ref{sec:ms}, respectively.

\subsection{Network Architecture}
\label{sec:na}
DNMOT consists of a backbone and an improved denoising Transformer architecture. In order to better integrate temporal information, we adopt multi-frame features. In this section, we will briefly introduce the details of our network architecture.

\textbf{Backbone.} We employ ResNet50~\cite{he2016deep} as our feature extractor, utilizing the output of the last three layers of the model. We apply another convolution layer with a kernel size of 3 and a stride of 2 for the final layer's output, and merge the features from the four scales to obtain the final feature vector $F\in \mathbb{R}^{B\times N_\text{feature}\times d}$. Here, $N_\text{feature}$ represents the total number of features across the four scales, and $d$ represents the dimension of the features.

\textbf{Multi-frame Features.} Following~\cite{wang2021end}, we merge the backbone features of the current and previous frames and input them together into the Transformer structure. This enables the model to directly compare the object's position between two frames.
%and learn that the object undergoes approximately uniform motion. 
To provide the model with 3D position embeddings, including time, we use trigonometric functions as 2D position embeddings in ViT~\cite{dosovitskiy2020image} and incorporate them into the Transformer~\cite{arnab2021vivit,bertasius2021space}. This helps the model to distinguish between two consecutive frames.

\textbf{Transformer.} Our main network architecture is based on the classic Transformer structure, which consists of an Encoder and a Decoder. The Encoder is made up of several consecutive layers, where each layer contains a multi-head self-attention module followed by a feed-forward network. Similarly, the Decoder also consists of several identical layers, and each layer includes a self-attention module, a cross-attention module, and a feed-forward network. 
%All the encoded features obtained from the Encoder are used as Key and Value in the Decoder. In section \ref{sec:qs}, we will introduce the generation process of the Query.

\subsection{Query Generation}
\label{sec:qs}

\begin{figure*}
    \centering
    \includegraphics[width=1.0\textwidth]{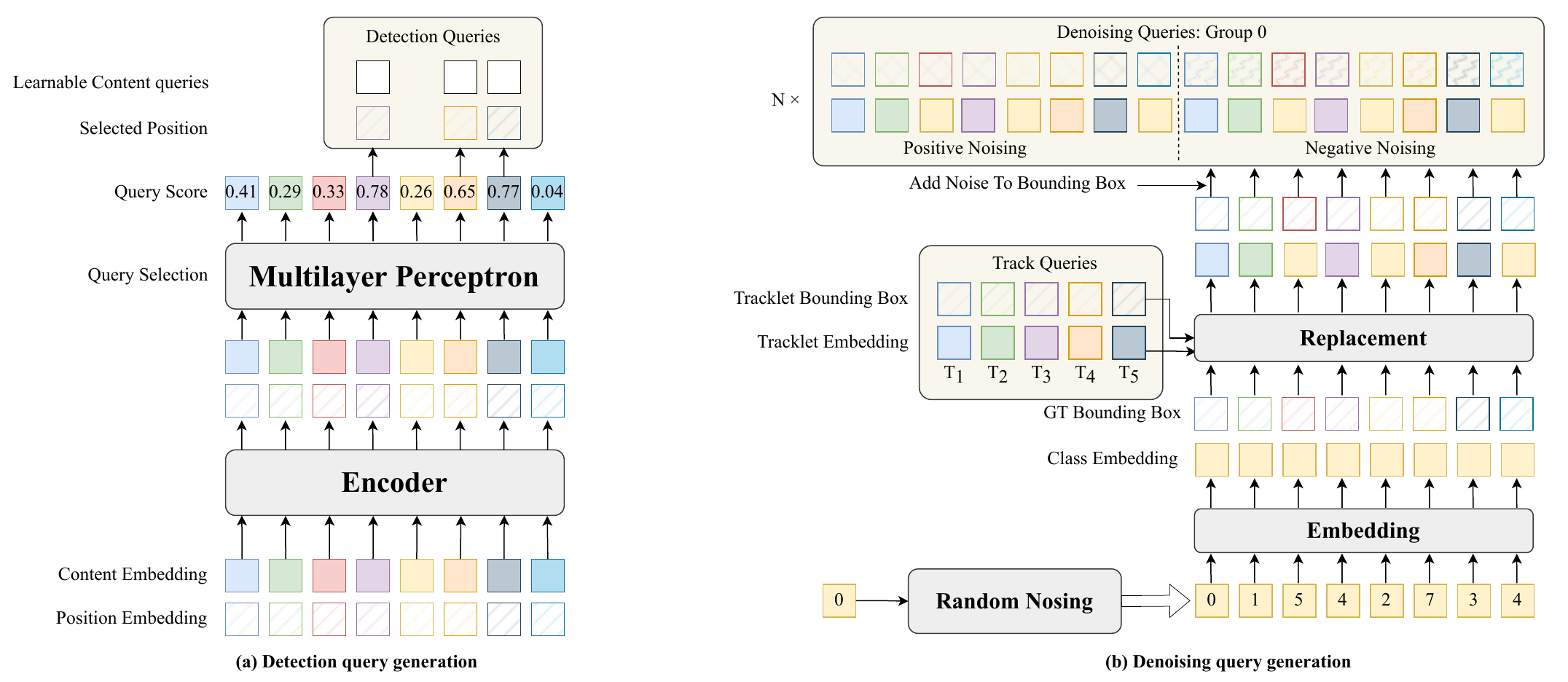}
    \caption{The generation process of queries in decoder. (a) illustrates how features extracted by backbone and encoder are filtered to provide location references for detection queries. (b) shows how to combine the existing active trajectories and ground truth information to generate the Denoising queries.}
    \label{figure4}
\end{figure*}

The overall generation process is illustrated in Figure \ref{figure4}. Figure \ref{figure4} (a) demonstrates how Query Selection is utilized to help the detection queries in more precisely locating the region of interest. Figure \ref{figure4} (b) illustrates how the final denoising queries are generated via track queries and Ground Truths.

\textbf{Detection Query.} As illustrated in Figure \ref{figure4}(a), the detection query is responsible for detecting new trajectories. This query needs to address two main difficulties: (1) it must detect all objects in the frame
%similar to the function of an object detector in the TBD paradigm
~\cite{bochkovskiy2020yolov4,he2017mask,liu2016ssd}; (2) 
%after interacting with the track query in the subsequent decoder's self-attention module, 
it must remove trajectories already activated in the detection results, as these trajectories are already represented by the track query. The detection queries are obtained by adding the content embedding and the position embedding. To help the queries find the region of interest faster, we use selected position embeddings instead of uniform position embeddings. Specifically, each output of encoder will generates a score by a MLP, which represents the probability of an object occurring at the location represented by the encoder output. The position embedding of the top $N_d$ queries, where $N_d$ is the number of detection queries and is typically set to 300, will be added with the same number of randomly initialized content queries to form the final detection query set. While we keep the learnable content queries to learn their own content embedding that helps to interact between different types of queries, we use the selected position embeddings to accelerate the detection process.

\textbf{Track Query.} Track query is used to continuously track all active trajectories, which are tracked and initialized in the previous frame. Track queries from the previous frame that generates high-confidence track results are considered active trajectories, and their content embedding and position embedding in the current frame are obtained from the final layer's output of the decoder and the generated object position, respectively. Additionally, new tracks with detection query results above the threshold $\lambda_D$ in the previous frame are treated as new tracks and added to the track query of the current frame for subsequent tracking.

\textbf{Denoising Query.} The denoising query is utilized to simulate the scenario of occlusion. To achieve this, we create $N$ groups of noise queries, where each group comprises the same number of positive noise as the objects present in the ground truth. Additionally, we generate an equal number of negative noise by utilizing a broader range of noise, inspired by the approach proposed in DINO~\cite{zhang2022dino}. Figure \ref{figure4}(b) depicts our generation process. Although our task involves only one category% of pedestrians
, we initially perturb the category numbers to a larger space with an uniform distribution, similar to DINO. The aim is not to differentiate between different categories. Instead, we adopt this method to yield a different mapping outcome for the category of pedestrians. This is because, in crowded scenes, occlusion tends to make pedestrians partially visible, thereby creating a significant difference between the acquired mapping vectors. Subsequently, we replace the embedding and the real object position % generated by the corresponding Ground Truth in the previous stage
with the existing content embedding and bounding box of the active trajectory. Finally, we employ various ranges of noise to perturb the position and map it to position embedding after perturbation. During the final loss calculation and back-propagation stage, we utilize the ground truth as supervision for positive noise and "no-object" as supervision information for negative noise.

We employ different strategies for generating positive noise depending on the presence of other objects around the object. A random noise vector $N=(n_x,n_y,n_w,n_h)$ is sampled, and all its elements are randomly chosen from a range between $-\lambda_r$ and $\lambda_r$ for each of the four values of the bounding box $B=(x,y,w,h)$. The resulting coordinate information is obtained after the addition of the noise vector:
\begin{equation}
    B_\text{new} =  B + B\odot N
\end{equation}

where $\odot$ represents element-wise product. When an object is close to another object (with IOU greater than threshold $\tau_c$), we introduce a conditional noise strategy, in which the final result is a weighted sum of the current object's bounding box and its neighboring object's bounding box:
\begin{equation}
    B_\text{new} = \lambda_c B + (1-\lambda_c)B_n
\end{equation}

where $B_n$ represents the bounding box of the neighboring object, and $\lambda_c$ is a conditional noise factor between 0 and 1. 
%We perform random noise generation with probability $p$ and conditional noise with probability $1-p$. 

\begin{figure}
  \centering
  \includegraphics[width=\linewidth]{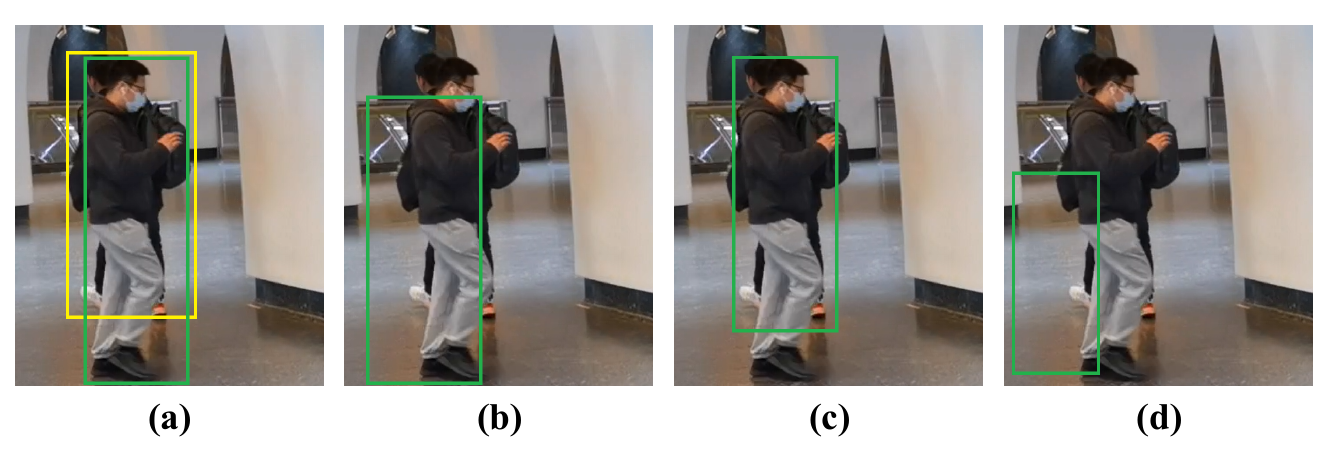}
  \caption{Visualization of different types of Noising. (a): ground truth of the object which we add noises to and its neighboring object. (b): with positive random noise. (c): with positive conditional noise. (d): with negative noise.}
  \label{figure5}
\end{figure}

Figure \ref{figure5} depicts our three noise addition methods. In (a), the bounding boxes of two neighboring objects are shown in green and yellow, respectively. (b), (c), and (d) demonstrate the results of positive random noising, positive conditional noising, and negative noising for the green object, respectively. Finally, the results of the three noise addition methods are merged to create a group of Denoising queries. The final Denoising query set comprises multiple groups of Denoising queries.

\begin{figure}
  \centering
  \includegraphics[width=0.7\linewidth]{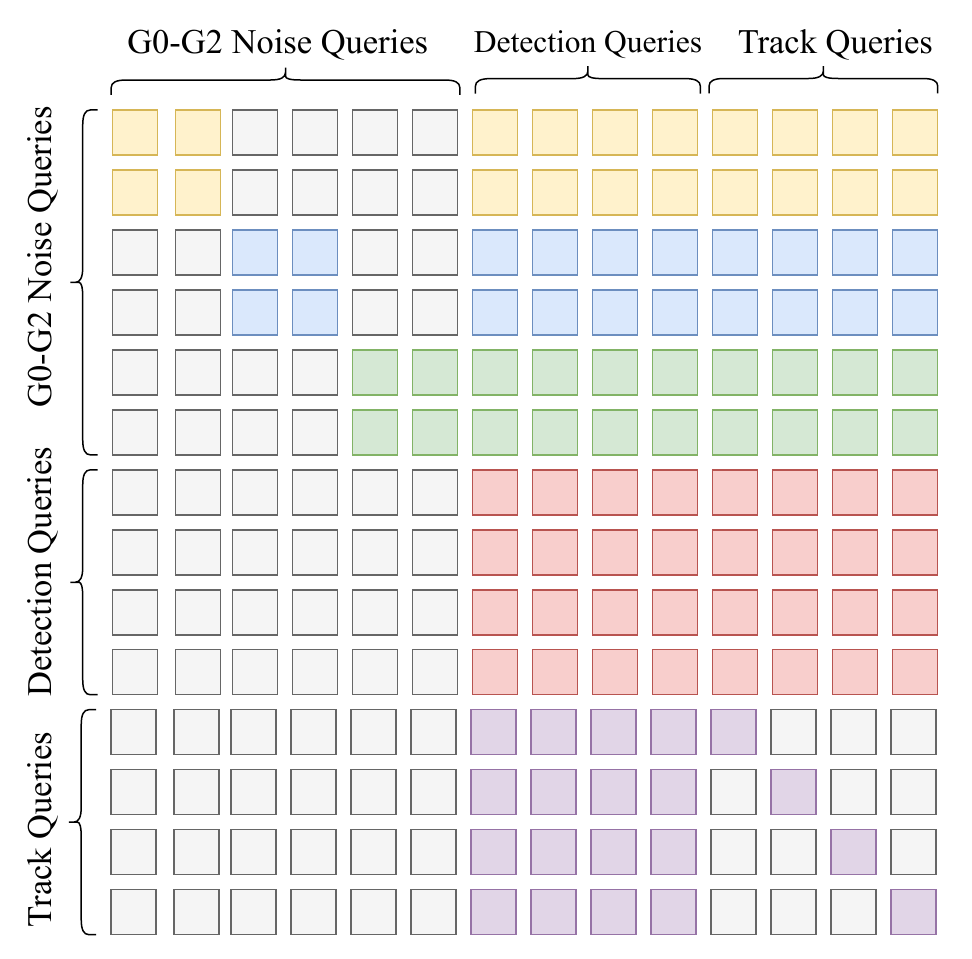}
  \caption{The cascaded mask used in the self-attention of the decoder layer. Different colors represent different types of queries or different groups of denoising queries. The gray parts indicate the invisibility between each other.}
  \label{figure6}
\end{figure}

\subsection{Cascaded Mask Self-attention}
\label{sec:ms}

In the Decoder, the query undergoes two processes: self-attention and cross-attention. During self-attention, queries interact with each other, potentially facilitating communication between queries carrying different information. In the subsequent cross-attention stage, the query interacts with all the keys and attends only to itself. All three types of queries require cross-attention to interact with features extracted earlier for fine-grained classification and regression tasks. However, the demands for the self-attention are different for each type of query.

For detection query, on the one hand, it needs to interact with other queries of the same type to ensure that no duplicate objects are detected. On the other hand, this type of query also needs to determine whether it is focusing on a new trajectory by interacting with the track queries. 

For track query, self-attention may have suppressed effects on other queries in its vicinity. 
%leading to a simple threshold filtering in the post-processing step to achieve a one-to-one correspondence between objects and outputs and reduce false positives. 
This mechanism is still needed for detection queries since their number is often larger than the actual number of new trajectories. However, in crowded scenes, trajectories often have tight spatial relationships with each other, and this mechanism often leads to the suppressed influence of occluded trajectory objects. On the other hand, the inter-relationship between trajectories, such as companions walking together, can help each other in the tracking process due to their similar speed and direction. Therefore, a mechanism that allows for interaction between trajectories without interfering with each other needs to be designed.

For Denoising query, it is a separate module. And it is important to ensure that different groups of denoising queries do not interfere with each other. This is because adding noise to different groups is a random process, and two groups of noise may potentially refer to each other. 
Therefore, it is necessary to ensure that any two groups of denoising queries are not visible to each other. 

We have implemented the analysis mentioned above using the mask self-attention mechanism in our model. The mask is represented as a $N_q\times N_q$ matrix, where $N_q$ is the number of queries. The $i$-th row corresponds to the invisibility mask for the $i$-th query in self-attention. The $j$-th location in this row is gray means that the $j$-th query is invisible to the $i$-th query during self-attention. 
%We use different colors to distinguish between different types of queries and different groups of denoising queries.
It should be noted that we use a progressive visibility scheme for track queries. In the initial few layers, typically the first half of the decoder layers, the track queries are fully visible to each other, and each track query interacts with other queries to extract coupled information between them. In the subsequent layers, typically the latter half of the layers, as shown in Figure \ref{figure6}, we introduce masks to limit each track query's interactions only with itself, thus preventing suppression by other queries.

\subsection{Loss Function}

Our model's optimization is a bipartite matching problem, and we utilize set prediction loss, as in other transformer-based approaches~\cite{carion2020end}. 

Each track query corresponds to either a ground truth trajectory or a `no-object' (which indicates the trajectory's termination in the current frame). For detection queries, we use DETR's bipartite matching mechanism to establish the correspondence between the model outputs and the ground truths. Concerning denoising queries, we supervise positive and negative noises with actual ground truth and `no-object' category, respectively. For classification score calculation, we employ focal loss \cite{lin2017focal}, and for bounding box regression, we use L1 and IOU loss \cite{rezatofighi2019generalized}. Our loss function is defined as:
\begin{equation}
    \mathcal{L}_\text{track} = \lambda_\text{focal}\mathcal{L}_\text{cls} + \lambda_\text{L1}\mathcal{L}_\text{bbox} + \lambda_\text{iou}\mathcal{L}_\text{iou}
    \label{e3}
\end{equation}
where $\lambda_\text{focal}$, $\lambda_\text{L1}$ and $\lambda_\text{iou}$ are the loss weights for balancing the focal loss, L1 loss and the IOU loss, respectively. In addition, similar to Deformable DETR~\cite{zhu2020deformable}, we add auxiliary losses after each decoder layer, and add extra intermediate losses after the query selection module, with the same components as for each decoder layer. Finally, our final loss function is defined as:
\begin{equation}
    \mathcal{L} = \mathcal{L}_\text{track} + \mathcal{L}_\text{aux} + \mathcal{L}_\text{inter}
\end{equation}
where $\mathcal{L}_\text{track}$ denotes the track loss in Equation \ref{e3}, $\mathcal{L}_\text{aux}$ denotes the auxiliary loss, and $\mathcal{L}_\text{inter}$ denotes the intermediate loss.

\section{Experiments}
In this section, we demonstrate the performance of our model on three public datasets, namely MOT17~\cite{milan2016mot16}, MOT20~\cite{dendorfer2020mot20}, and DanceTrack~\cite{sun2022dancetrack}. In addition, we conduct ablation experiments to verify the effectiveness of our modules.

\subsection{Datasets and Metrics}

\textbf{Datasets.} We evaluated our model's performance on three publicly available datasets: MOT17~\cite{milan2016mot16}, MOT20~\cite{dendorfer2020mot20}, and Dance Track~\cite{sun2022dancetrack} to ensure fair comparison. MOT17 consists of 7 training and 7 testing sequences, each with detection results from three existing detectors~\cite{ren2015faster,felzenszwalb2009object,yang2016exploit} for evaluating the association performance of methods. As our model is an end-to-end approach, we only conducted experiments on the private track. MOT20 includes 4 training and 4 testing sequences with more objects in the scene compared to MOT17. DanceTrack is a recent multi-object tracking dataset with 100 dance sequences, including 40 training sequences, 25 validation sequences, and 35 testing sequences.

\textbf{Validation set.} We conducted our ablation studies on MOT17. Due to the restrictions on the number of submissions for the test set in MOT Challenge~\cite{milan2016mot16,dendorfer2020mot20}, we sampled half of each training sequence as our validation set followed by~\cite{zhang2021fairmot}.

\textbf{Metrics.} We use CLEAR~\cite{bernardin2008evaluating} MOT metrics and HOTA~\cite{luiten2021hota} as our evaluation metrics.

\begin{table}[]
    \centering
    \caption{Performance comparison between DNMOT and existing methods on the MOT17 dataset under the private detection protocols.}
    \scalebox{0.77}{
    \begin{tabular}{lcccccc}
    \toprule    Method & MOTA $\uparrow$ & IDF1 $\uparrow$ & HOTA $\uparrow$ & FP $\downarrow$ & FN $\downarrow$ & ID.Sw $\downarrow$\\
    \midrule
    TraDeS~\cite{wu2021track} & 69.1 & 63.9 & 52.7 & 20892 & 150060 & 3555\\
    
    FairMOT~\cite{zhang2021fairmot} & 73.7 & 72.3 & 59.3 & 27507 & 117477 & 3303\\
    GTR~\cite{zhou2022global} & 75.3 & 71.5 & 59.1 & 26793 & 109854 & 2859\\
    CorrTracker~\cite{wang2021multiple} & 76.5 & 73.6 & 60.7 & 29808 & 99510 & 3396\\
    OC-SORT~\cite{cao2022observation} & 78.0 & 77.5 & 63.2 & 15100 & 108000 & 1950\\
    MOTRv2~\cite{zhang2022motrv2} & 78.6 & 75.0 & 62.0 & 23409 & 94797 & 2619\\
    GHOST~\cite{seidenschwarz2022simple} & 78.7 & 77.1 & 62.8 & - & - & 2325\\
    ByteTrack ~\cite{zhang2022bytetrack} & 80.3 & 77.3 & 63.1 & 25491 & 83721 & 2196\\
    BoT-SORT~\cite{aharon2022bot} & 80.6 & 79.5 & 64.6 & 22524 & 85398 & 1257\\
    C-BIOU Tracker~\cite{yang2023hard} & 81.1 & 79.7 & 64.1 & 23136 & 82011 & 1455\\
    \midrule
    TransCenter~\cite{xu2021transcenter} & 73.2 & 62.2 & 54.5 & 23112 & 123738 & 3663\\
    TransTrack~\cite{sun2020transtrack} & 74.5 & 63.9 & 54.1 & 28323 & 112137 & 3663\\
     \rowcolor{mygray}MeMOT~\cite{cai2022memot} & 72.5& \textbf{69.0} & 56.9 & 37221 & 115248 & 2724\\
     \rowcolor{mygray}MOTR~\cite{zeng2022motr} & 73.4 & 68.6 & 57.8 & - & - & \textbf{2439}\\
     \rowcolor{mygray}TrackFormer~\cite{meinhardt2022trackformer} & 74.1 & 68.0 & 57.3 & 34602 & \textbf{108777} & 2829\\
     \rowcolor{mygray}DNMOT(Ours) & \textbf{75.6} & 68.1 & \textbf{58.0} & \textbf{24960} & 110064 & 2529\\
    \bottomrule
    \end{tabular}}

    \label{tab1}
\end{table}

\subsection{Implementation Details}

We utilized PyTorch~\cite{paszke2019pytorch} to develop our model and carried out experiments on 8 NVIDIA 3090Ti GPUs. The image was resized to a minimum size of 800, and we employed data augmentation, such as random flipping and cropping. %Following Trackformer~\cite{meinhardt2022trackformer}, we pre-trained our model on CrowdHuman for 80 epochs.

\textbf{Hyperparameters.} Our model was pre-trained on Crowdhuman~\cite{shao2018crowdhuman} for 80 epochs, followed by 40, 50, and 20 epochs of training on MOT17~\cite{milan2016mot16}, MOT20~\cite{dendorfer2020mot20}, and DanceTrack~\cite{sun2022dancetrack}, respectively. The initial learning rate was $2\times 10^{-4}$ and was decayed after the 10 epochs. We used the AdamW~\cite{loshchilov2017decoupled} optimizer. Our Transformer structure consisted of 6 encoder layers and 6 decoder layers, with 8 heads in the attention mechanism. After conducting ablation experiments, we set the number of detection queries to 300 and determined the number of denoising query groups based on the number of objects in the ground truth, with a total denoising quantity of no more than 200. The batch size was initially set to 2, but it was adjusted to 1 when using the multi-frame strategy due to memory limitations. We set $\lambda_{focal}$, $\lambda_{L1}$, and $\lambda_{iou}$ to 1, 5, and 2, respectively.

\subsection{Benchmark Results}
In this section, we report the results of our experiments on three datasets: MOT17~\cite{milan2016mot16}, MOT20~\cite{dendorfer2020mot20}, and DanceTrack~\cite{sun2022dancetrack}. Our experiments demonstrate that our method achieves state-of-the-art or comparable results under the MOTA and HOTA metrics among Transformer-based end-to-end multi-object trackers that do not rely on external detectors (e.g., YOLOX~\cite{ge2021yolox}). Trackers with gray background do not require additional modules for inference, and the best results among them are marked in bold.

\textbf{MOT17.} In Table~\ref{tab1}, we present the results of our experiments on MOT17~\cite{milan2016mot16}. Our DNMOT achieves the best performance among all methods that do not require additional modules for inference, with a MOTA of 75.6 and a HOTA of 58.0. It is worth mentioning that we only use the CrowdHuman~\cite{shao2018crowdhuman} dataset for pre-training, without any other datasets~\cite{dollar2009pedestrian,dollar2011pedestrian,zhang2017citypersons,xiao2017joint,ess2008mobile,zheng2017person}, yet we achieve comparable or even better results. Our model successfully tracks many invisible objects, with a low number of false positives and false negatives.

\textbf{MOT20.} We present our experimental results on the MOT20 dataset \cite{dendorfer2020mot20} in Table~\ref{tab2}. Our DNMOT achieved 70.5 MOTA, 73.2 IDF1, and 58.6 HOTA, which are the best results among similar methods. Compared to MOT17, MOT20 has a more crowded distribution of objects, and our model's high robustness to occlusion leads to a greater improvement on MOT20 (+1.9 in MOTA, +7.1 in IDF1 and +3.9 in HOTA) than on MOT17 (+1.5 in MOTA, -0.9 in IDF1 and +0.2 in HOTA).

\begin{table}[]
    \centering
    \caption{Performance comparison between DNMOT and existing methods on the MOT20 dataset.}
    \scalebox{0.77}{
    \begin{tabular}{lcccccc}
    \toprule    Method & MOTA $\uparrow$ & IDF1 $\uparrow$ & HOTA $\uparrow$ & FP $\downarrow$ & FN $\downarrow$ & ID.Sw $\downarrow$\\
    \midrule
    FairMOT~\cite{zhang2021fairmot} & 61.8 & 67.3 & 54.6 & 103440 & 88901 & 5243\\
    CorrTracker~\cite{wang2021multiple} & 65.2 & 69.1 & - & 79429 & 95855 & 5183\\
    OC-SORT~\cite{cao2022observation} & 75.5 & 75.9 & 62.1 & 18000 & 108000 & 913\\
    ByteTrack~\cite{zhang2022bytetrack} & 77.8 & 75.2 & 61.3 & 26249 & 87594 & 1223\\
    GHOST~\cite{seidenschwarz2022simple} & 73.7 & 75.2 & 61.2 & - & - & 1264\\
    BoT-SORT~\cite{aharon2022bot} & 77.8 & 77.5 & 63.3 & 24638 & 88863 & 1257\\
    \midrule
    TransTrack~\cite{sun2020transtrack} & 64.5 & 59.2 & 48.9 & 28566 & 151377 & 3565\\
    TransCenter~\cite{xu2021transcenter} & 67.7 & 58.7 & 43.5 & 56435 & 107163 & 3759\\
     \rowcolor{mygray}MeMOT~\cite{cai2022memot} & 63.7& 66.1 & 54.1 & 47882 & 137983 & 1938\\
     \rowcolor{mygray}TrackFormer~\cite{meinhardt2022trackformer} & 68.6 & 65.7 & 54.7 & \textbf{20348} & 140373 & 1532\\
     \rowcolor{mygray}DNMOT(Ours) & \textbf{70.5} & \textbf{73.2} & \textbf{58.6} & 29314 & \textbf{122252} & \textbf{987}\\
    \bottomrule
    \end{tabular}}

    \label{tab2}
\end{table}

\begin{table}[]
    \centering
    \caption{Performance comparison between DNMOT and existing methods on the DanceTrack test set.}
    \scalebox{0.8}{
    \begin{tabular}{lccc}
    \toprule    Method & MOTA $\uparrow$ & IDF1 $\uparrow$ & HOTA $\uparrow$\\
    \midrule
    OC-SORT~\cite{cao2022observation} & 75.5 & 75.9 & 62.1\\
    TraDeS~\cite{wu2021track} & 86.2 & 41.2 & 43.3\\
    ByteTrack~\cite{zhang2022bytetrack} & 89.6 & 53.9 & 47.7 \\
    GHOST~\cite{seidenschwarz2022simple} & 91.3 & 57.7 & 56.7 \\
    MOTRv2~\cite{zhang2022motrv2} & 92.1 & 76.0 & 73.4\\
    \midrule
    
     \rowcolor{mygray}MOTR~\cite{zeng2022motr} & 79.7& \text{54.2} & 51.5 \\
     \rowcolor{mygray}DNMOT(Ours) & \textbf{89.1} & 49.7 & \textbf{53.5} \\
    \bottomrule
    \end{tabular}}

    \label{tab3}
\end{table}

\textbf{DanceTrack.} We conduct experiments on the recently proposed DanceTrack dataset \cite{sun2022dancetrack}, which focuses on evaluating the performance of the tracker association stage using the HOTA evaluation metric~\cite{luiten2021hota}. In Table~\ref{tab3}, we present our results, which show that our approach achieves a HOTA score of 53.5, outperforming comparable methods by 2.0 points. Furthermore, our approach achieves a MOTA score of 89.1 and an IDF1 score of 49.7. 

\subsection{Ablation Study}

In this section, we conduct ablation experiments to demonstrate the effectiveness of our proposed method and discuss the model hyperparameters. 

\textbf{DNMOT Components.} Table \ref{tabadd} shows the impact of integrating different components. The Baseline is Deformable DETR ~\cite{zhu2020deformable}, and we add track query ~\cite{meinhardt2022trackformer} to make baseline applicable to MOT tasks. Integrating our components into the baseline can gradually improve overall performance. The application of denoising queries gives the model stronger performance, improving MOTA by 3.8 and IDF1 by 5.6 over baseline, as well as reducing ID Switch by 77\%. By adding multi-frame features, the model can learn the motion patterns of objects in two frames, further improving MOTA and IDF1 and reducing the number of ID switches. When using Query Selection Module, there are extra 0.4 and 0.6 improvements in MOTA and IDF1, respectively. Finally, after adding the cascaded mask, we obtain 75.4 MOTA, 71.0 IDF1 and 477 ID switches, which is a significant improvement over the original Baseline (5.1 MOTA, 8.0 IDF1 improvement and 87\% ID switch reduction).
\begin{table}[]
    \centering
    \caption{The effect of our contributions. Baseline is Deformable DETR with track queries added.}
    \scalebox{0.9}{
    \begin{tabular}{l|ccc}
    \toprule 
    Method & MOTA$\uparrow$ & IDF1$\uparrow$ & ID.Sw$\downarrow$ \\
    \midrule
    Baseline & 70.3 & 63.0 & 3264\\
    +Denoising query & 74.1 & 68.6 & 774\\
    +Multi-frame feature & 74.4 & 69.5 & 699\\
    +Query Selection & 74.8 & 70.1 & 496\\
    +Cascaded Mask & \textbf{75.4} & \textbf{71.0} & \textbf{477}\\
    \bottomrule
    \end{tabular}
    }
    \label{tabadd}
\end{table}
\begin{table}[]
    \centering
    \caption{The results of the experiments on different types of noise. We denote the different types of noise as follows: PRN: Positive Random Noises, NN: Negative Noises, PCN: Positive Conditional Noises.}
    \scalebox{0.9}{
    \begin{tabular}{ccc|ccc}
    \toprule 
    PRN & NN & PCN & MOTA$\uparrow$ & IDF1$\uparrow$ & ID.Sw$\downarrow$ \\
    \midrule
    &  &  & 70.3 & 63.0 & 3264\\
    \checkmark &  &  & 72.9 & 64.6 & 3636\\
    \checkmark & \checkmark &  & 73.2 & 65.8 & 1112\\
    \checkmark & \checkmark & \checkmark & \textbf{74.1} & \textbf{68.6} & \textbf{774}\\
    \bottomrule
    \end{tabular}}
    \label{tab4}
\end{table}

\textbf{Noising Type.} We evaluate the effects of three noise types and present the results in Table \ref{tab4}. The denoising part is removed as our baseline for ablation study. Adding positive random noises (second row in the table) improves the bounding box prediction results, resulting in a significant increase in MOTA(+2.6).  The more accurate position output leads to better position embedding for the next frame's track query, reducing the number of ID Switches. The addition of negative noises (third row in the table) resulted in the rejection of proposals that are farther away, further improving MOTA (+0.3) and IDF1 (+1.2) while reducing the number of ID switches. The introduction of positive conditional noises results in the best performance on the validation set, achieving 74.1 MOTA, 68.6 IDF1, and 774 ID switch numbers.

\textbf{Cascaded Mask.} We conduct experiments to demonstrate the effectiveness of our mask module and present our experimental results in Table \ref{tab5}. We use Trackformer~\cite{meinhardt2022trackformer} as our baseline and conducted ablation experiments on it (first row in the table). We modify the self-attention part in the decoder while keeping other parameters constant. We try three modifications: processing track queries separately using an MLP (second row in the table), adding a mask to all decoder layers for track queries (third row in the table), and gradually adding a mask to track queries using a cascaded approach (fourth row in the table).

We use an MLP with only linear and ReLU layers to process the track queries. However, our experiment shows a slight drop in performance when using an MLP to process the track queries separately. We believe this is due to two reasons. First, there are logical relationships between track queries that can aid in easier tracking by linking them together. The use of an MLP breaks this connection. Second, using attention and MLP to process two types of queries separately may not result in the same feature space, making it challenging for subsequent cross-attention and thus affecting the model's optimization.

We further utilize the mask mechanism to ensure that the outputs of all query self-attention belong to the same feature space. However, using masks for all track queries throughout the decoder results in a decrease in metrics, suggesting that the interaction between tracks is still meaningful. To address this issue, we adopt a gradually added mask mechanism where masks are not added to track queries in the early decoder layers, allowing track queries to interact with each other. In later layers, masks are added so that track queries could focus on their own objects without interference. Our experiments demonstrate that this Cascaded Mask module  is highly effective and helps our model achieve the best results.

\begin{table}[]
    \centering
    \caption{Experimental results with different mask methods.}
    \scalebox{0.9}{
    \begin{tabular}{l|ccc}
    \toprule 
    Method & MOTA$\uparrow$ & IDF1$\uparrow$ & ID.Sw$\downarrow$ \\
    \midrule
    Baseline~\cite{meinhardt2022trackformer} & 74.2 & \textbf{71.8} & 1449\\
    MLP & 74.1 & 71.5 & 1533\\
    Full Mask & 69.4 & 60.2 & 1086\\
    Cascaded Mask & \textbf{74.5} & 68.6 & \textbf{838}\\
    \bottomrule
    \end{tabular}
    }
    \label{tab5}
\end{table}

% \begin{table}[]
%     \centering
%     \caption{Experimental results with selections of different number of queries and different epochs.}
%     \scalebox{0.9}{
%     \begin{tabular}{cc|ccc}
%     \toprule 
%     $N_q$ & epochs & MOTA$\uparrow$ & IDF1$\uparrow$ & ID.Sw$\downarrow$ \\
%     \midrule
%     150 & 40 & 73.5 & 67.5 & 1177\\
%     150 & 80 & 74.6 & 68.4 & 883\\
%     300 & 40 & 73.8 & 66.8 & 1039\\
%     300 & 80 & \textbf{74.8} & \textbf{70.1} & \textbf{496}\\
%     \bottomrule
%     \end{tabular}
%     }
%     \label{tab6}
% \end{table}

% \textbf{Number of Detection Queries.} We conduct experiments to investigate the effect of different numbers of detection queries on the final results and present our findings in Table \ref{tab6}. As the maximum number of objects in a frame in MOT17 is 52, the number of queries should be much larger in practice. Therefore, we choose 150 and 300 queries and train the model for 40 and 80 epochs for each selection, respectively. Our experiments show that despite the number of detection queries being already much larger than the number of new trajectories to be detected, increasing the number of queries still leads to a moderate performance improvement in the final results. Moreover, as the number of queries increases, the model becomes less sensitive to the number of epochs. 

\begin{table}[]
    \centering
    \caption{The experimental results of choosing the hpyerparameter $\lambda_d$ and $\lambda_t$.} %We used Trackformer as the baseline and modified the self-attention in the decoder to validate our different mask strategies.}
    \scalebox{0.9}{
    \begin{tabular}{b{0.7cm} b{0.7cm}|ccc}
    \toprule 
    \centering$\lambda_d$ & \centering$\lambda_t$ & MOTA$\uparrow$ & IDF1$\uparrow$ & ID.Sw$\downarrow$ \\
    \midrule
    \centering0.4 & \centering0.3 & \textbf{73.9} & \textbf{66.7} & \textbf{1005}\\
    \centering0.4                     & \centering0.4 & 73.6 & 66.6 & 1042\\
    \centering0.4                     & \centering0.5 & 72.8 & 66.1 & 1167\\
    \midrule
    \centering0.3 & \centering0.4 & \textbf{73.7} & 65.9 & 1298\\
    \centering0.4 &  \centering0.4 & 73.6 & \textbf{66.6} & 1042\\
    \centering0.5 &  \centering0.4 & 72.9 & 66.5 & \textbf{863}\\
    \bottomrule
    \end{tabular}
    }
    \label{tab7}
\end{table}
% \centering\multirow{3}{*}{0.4}
\textbf{Threshold Selection.} We investigate the impact of different threshold choices for different queries on the final results. The experimental results are shown in Table \ref{tab7}. For track queries, the best performance in terms of MOTA and IDF1 is achieved with a value of 0.3 for parameter $\lambda_t$. Conversely, for detection queries, the performance of the model is not sensitive to changes in parameter $\lambda_d$.

\subsection{Discussion}
% In the supplementary material, we counted the occlusion situations in different datasets, which were quantified by calculating the number of object pairs with an IOU~\cite{rezatofighi2019generalized} greater than a certain value in each frame. By analyzing the results, we found that MOT17 is not very crowded, with only three target pairs with an IOU greater than 0.6. On the other hand, MOT20 is much more crowded, but this is only for IOU. In fact, most sequences in MOT20 are shot from a top-down perspective, and although people are very close to each other, they do not obstruct each other. Therefore, compared to improvements in accuracy, our method has greater value in application.

%In the supplementary material, we conduct an analysis of occlusion situations in different datasets by calculating the number of object pairs with an IOU score larger than a certain value in each frame, as quantified by \cite{rezatofighi2019generalized}. The results show that
In terms of IOU scores between objects, MOT17 is not heavily crowded and MOT20 is more crowded. However, in MOT20, sequences are often shot from a top-down perspective, and although people are close to each other, they do not obstruct each other. Therefore, compared to improvements in MOTA and HOTA, our method has greater value in application.

\section{Conclusion}

In this work, we propose DNMOT, an end-to-end trainable multi-object tracker based on the Transformer architecture. DNMOT adds noises during training and learns to denoising, resulting in stronger robustness in severe occluded scenes. Our method achieves state-of-the-art performance among all methods that do not require any additional modules during inference. 

We hope those methods will foster future work for multi-object tracking with severe occlusions.
\begin{sloppypar}
\section*{Acknowledgements}
This work was supported in part by the National Natural Science Foundation of China (No.62176060), STCSM project (No.22511105000), Shanghai Municipal Science and Technology Major Project (No.2021SHZDZX0103), and the Program for Professor of Special Appointment (Eastern Scholar) at Shanghai Institutions of Higher Learning.
\newpage
\balance
\bibliographystyle{ACM-Reference-Format}
\bibliography{sample-base}

\appendix
\onecolumn

\section{details of network architecture in ablation experiments}

\begin{figure}
  \centering
  \includegraphics[width=\linewidth]{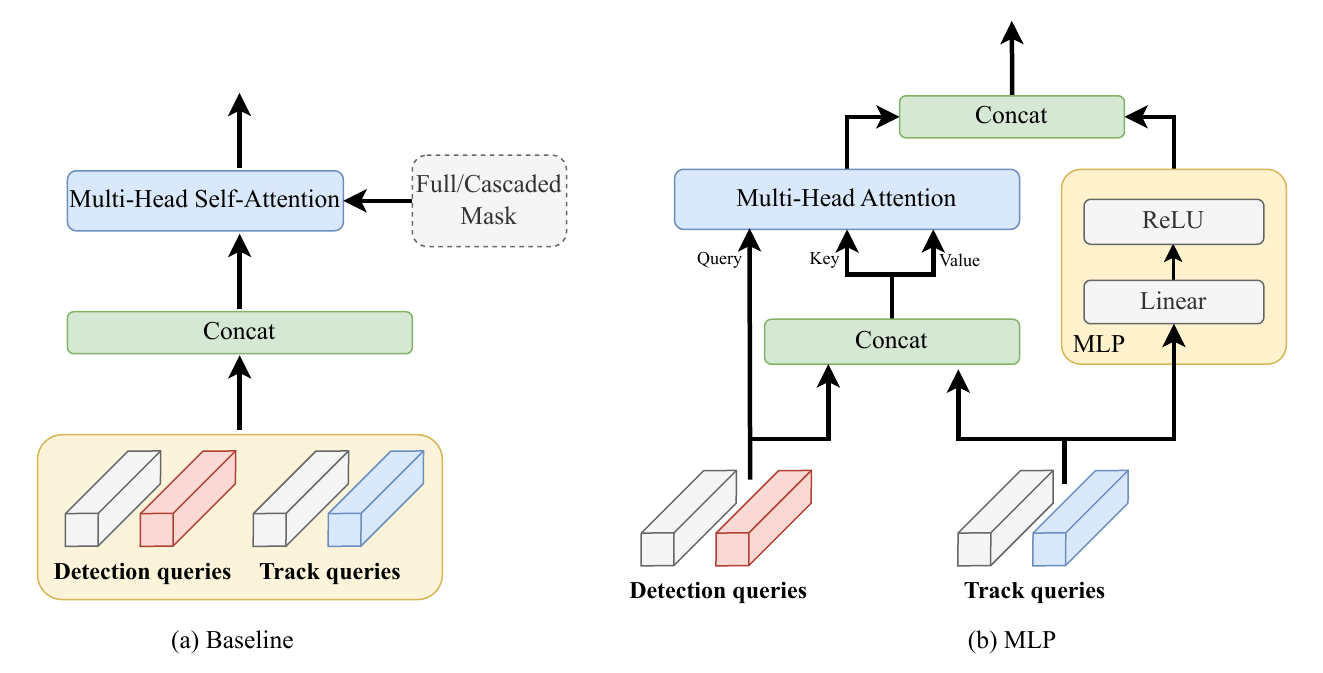}
  \caption{Network architecture in the ablation experiment of "Cascaded Mask" in the main text. (a) illustrates our baseline (without "Full/Cascaded Mask"), which is used in most Transformer-based models, and we add different masks innovatively. (b) illustrates the architecture of MLP we performed in the experiment. We utilize the same processing method for denoising queries across all experiments, so we omit it from the figure.}
  \label{figs1}
\end{figure}

In this section, we provide the details of the network architecture used in the ablation experiment of ``Cascaded Mask'' in Figure \ref{figs1}. We use the ablation experiment to explore the interaction between different types of queries. Because Denoising queries have the same processing method in all the experiments, we omit them in the figure.

In the previous methods~\cite{meinhardt2022trackformer,zeng2022motr}, as in Figure \ref{figs1}(a) (without Full/Cascaded Mask), different types of queries are concatenated and passed to Self-Attention module to interact with each other. Considering the track queries are invisible to each other, we design an MLP to replace the original Self-Attention module. To process detection queries, all the queries are concatenated like previous methods as Key and Value, while only detection queries are used as Query. On the other hand, for the track queries, we use a MLP network, which contains a linear layer and an activation function. Finally, track queries and detection queries are concatenated and passed to the next stage.

As indicated in Table \textcolor{red}{6} in the main text, we finally use the mask to coordinate the interaction between different types of queries (Figure \ref{figs1}(a) with Full/Cascaded Mask).

\section{Algorithm of Generation of Denoising queries}

We provide the generation process of Denoising queries in Algorithm \ref{alg1}, as a more detailed description of Figure \textcolor{red}{4}(b) in the main text.

\begin{algorithm}[t]

\caption{The generation process of denoising queries}\label{algorithm}

\KwIn{Bounding boxes in ground truth $B = \left \{ B_{1},B_{2},...,B_{N}\right \} $\; 
Active trajectories $T = \left \{ T_{1},T_{2},...,T_{N}\right \} $\;
Correspondence between trajectories and ground truth $M$;\\
Category perturbation range C, conditional noise threshold $\lambda_c$, conditional noise probability $p$;\\
positive noise range $\lambda_r$, negative noise range $\lambda_n$;
}
\KwOut{$N_g$ groups of denoising queries}
Initialize group number of denoising queries: $N_g = 200 // B.length$  \\
\label{alg1}
% \KwOut{Best-Trained CNN Model $\phi \left ( \theta^{\ast}  \right )$.}
\For{group = 1 to $N_g$}{
    Randomly take $B.length$ category IDs from the range $[0, C]$  to form class set $S_a$;\\
    Map IDs in $S_a$ to class embeddings;\\
    Assign a bounding box in $B$ to each class embedding;\\
    Replace the corresponding class embedding with the tacklet embedding according to $M$;\\
    \For{each bounding box and class embedding pair}{
        Calculate the IOU scores between the bounding box and others;\\
        \uIf{At least an IOU score greater than $\lambda_c$}{
            \uIf{Random number $\le p$}{
                Random select a bounding box which IOU score greater than $\lambda_c$;\\
                Generate a Noise position via Eq \textcolor{red}{2} in the main text;
            }
            \uElse{
                Sample four noise factors $N=(n_x,n_y,n_w,n_h)$ from range $[-\lambda_r, \lambda_r]$;\\
                Generate a Noise position via Eq \textcolor{red}{1} in the main text;
            }
            Add the result to positive set $S_p$;\\
        }
        \uElse{
            Sample four noise factors $N=(n_x,n_y,n_w,n_h)$ from range $[-\lambda_r, \lambda_r]$;\\
            Generate a Noise position via Eq \textcolor{red}{1} in the main text;
            \\Add the result to $S_p$;
        }
        Sample four noise factors $N=(n_x,n_y,n_w,n_h)$ from range $[-\lambda_n, \lambda_n]$;\\
        Generate a Noise position via Eq \textcolor{red}{1} in the main text;
        add the result to negative set $S_n$;
    }
    Concatenate $S_p$ and $S_n$ as a group of denoising queries.
}

\end{algorithm}

\section{analysis of datasets}
In this section, we analyze the degree of crowdedness of the MOT17, MOT20 and DanceTrack dataset. To quantify this, we count the number of frames and pedestrians in training set of each dataset, and then we calculate the number of pairs with IOU scores greater than specific thresholds for each sequence. This allows us to obtain the average number of crowded pairs and pedestrians for each dataset, which characterizes their density levels.

For MOT17, our statistical results are demonstrated in Table \ref{tabs2}. %There are \textbf{204701} pedestrians in total, located in a total of 5316 frames, so there are around \textbf{38} pedestrians in a frame on average, while there are only around \textbf{2.5} pairs of pedestrians can have an IOU score greater than 0.4.
MOT17 contains \textbf{204,701} pedestrians spread across 5,316 frames, giving an average of \textbf{38} pedestrians per frame. Additionally, on average, there are only \textbf{2.5} pairs of pedestrians with an Intersection over Union (IOU) score greater than 0.4, indicating relatively low crowdedness in the dataset.

For MOT20, our statistical results are demonstrated in Table \ref{tabs4}. %there are \textbf{1134614} pedestrians in total, located in a total of 8931 frames, so there are around \textbf{127} pedestrians in a frame on average. There are about \textbf{20} pairs of pedestrians can have a IOU score more than 0.4. 
MOT20 contains a total of \textbf{1,134,614} pedestrians distributed over 8,931 frames, yielding an average of \textbf{127} pedestrians per frame. On average, about \textbf{20} pairs of pedestrians have an IOU score greater than 0.4. So MOT20 is more crowded in both the average number of people per frame and the average number of crowded pairs, and our model’s high robustness to occlusion leads to a greater improvement on MOT20 (+1.9 in MOTA, +7.1 in IDF1 and +3.9 in HOTA) than on MOT17 (+1.5 in MOTA, -0.9 in IDF1 and +0.2 in HOTA). 

For DanceTrack, our statistical results are demonstrated in Table \ref{tabs3}. There are \textbf{348930} pedestrians in a total of 41796 frames, which means there are around \textbf{8} pedestrians in a frame on average, while there is around \textbf{1} pair of pedestrians can have a IOU score more than 0.4. Compared to MOT17, DanceTrack has a more sparse object distribution, but a similar number of crowded pairs. %Coupled with the fact that the objects in this dataset have a more similar dress, the performance (+2.0 in HOTA) on this dataset better illustrates the effectiveness of our model.
Moreover, the objects in DanceTrack have similar dress, which makes it more challenging to distinguish and track them. The significant improvement in performance (+2.0 in HOTA) achieved by our model on DanceTrack further validates its effectiveness in handling complex scenarios.
\begin{table}[]
    \centering
    \caption{Degree of crowdedness in train set of MOT17 dataset.}
    \scalebox{1.0}{
    \begin{tabular}{lccccccc}
    \toprule    Sequence & 0.1-0.4 & 0.4-0.5 & 0.5-0.6 & 0.6-0.7 & 0.7-0.8 & 0.8-0.9 & \textgreater 0.9\\
    \midrule
    MOT17-02 & 9648 & 1113 & 1126 & 1104 & 545 & 187 & 7\\
    MOT17-04 & 16696 & 2292 & 1387 & 1068 & 350 & 75 & 0\\
    MOT17-05 & 2143 & 202 & 114 & 61 & 33 & 8 & 4\\
    MOT17-09 & 2496 & 344 & 380 & 213 & 136 & 33 & 0\\
    MOT17-10 & 3382 & 493 & 318 & 190 & 98 & 26 & 1\\
    MOT17-11 & 2770 & 222 & 48 & 36 & 6 & 1 & 0\\
    MOT17-13 & 3222 & 376 & 278 & 231 & 167 & 33 & 8\\
    \midrule
    Overall & 40357 & 5042 & 3651 & 2903 & 1302 & 367 & 20\\
    Average per frame & 7.5916 & 0.9484 & 0.6868 & 0.5461 & 0.2449 & 0.0690 & 0.0038\\
    \bottomrule
    \end{tabular}}
    \label{tabs2}
\end{table}

\begin{table}[]
    \centering
    \caption{Degree of crowdedness in train set of MOT20 dataset. MOT20 has more pedestrians and pairs that have an IOU score greater than 0.4.}
    \scalebox{1.0}{
    \begin{tabular}{lccccccc}
    \toprule    Sequence & 0.1-0.4 & 0.4-0.5 & 0.5-0.6 & 0.6-0.7 & 0.7-0.8 & 0.8-0.9 & \textgreater 0.9\\
    \midrule
    MOT20-01 & 15706 & 2157 & 1476 & 704 & 254 & 53 & 0\\
    MOT20-02 & 164874 & 19619 & 11632 & 6141 & 2695 & 909 & 94\\
    MOT20-03 & 224287 & 15828 & 7300 & 2759 & 908 & 281 & 19\\
    MOT20-05 & 669205 & 60809 & 31285 & 12185 & 2963 & 883 & 199\\
    \midrule
    Overall & 1074072 & 98413 & 51693 & 21789 & 6820 & 2126 & 312\\
    Average per frame & 128.0029 & 11.7284 & 6.1605 & 2.5967 & 0.8128 & 0.2537 & 0.0372\\
    \bottomrule
    \end{tabular}}
    \label{tabs4}
\end{table}

\begin{table}[]
    \centering
    \caption{Degree of crowdedness in train set of DanceTrack dataset. DanceTrack has a more sparse people distribution but similar number of crowded pairs. And people in DanceTrack have more similar dress.}
    \scalebox{1.0}{
    \begin{tabular}{lccccccc}
    \toprule    Sequence & 0.1-0.4 & 0.4-0.5 & 0.5-0.6 & 0.6-0.7 & 0.7-0.8 & 0.8-0.9 & \textgreater 0.9\\
    \midrule
    dancetrack0001 & 1162 & 271 & 120 & 35 & 15 & 1 & 0\\
    dancetrack0002 & 1938 & 218 & 196 & 162 & 47 & 7 & 0\\
    dancetrack0006 & 4160 & 704 & 545 & 336 & 119 & 23 & 2\\
    dancetrack0008 & 1901 & 271 & 231 & 155 & 116 & 32 & 1\\
    dancetrack0012 & 8149 & 1343 & 894 & 475 & 199 & 37 & 3\\
    dancetrack0015 & 5496 & 363 & 113 & 27 & 5 & 0 & 0\\
    dancetrack0016 & 4394 & 581 & 248 & 102 & 36 & 1 & 0\\
    dancetrack0020 & 19402 & 2506 & 1071 & 574 & 209 & 40 & 6\\
    dancetrack0023 & 4295 & 878 & 682 & 355 & 124 & 38 & 2\\
    dancetrack0024 & 1151 & 152 & 70 & 40 & 11 & 1 & 0\\
    dancetrack0027 & 1816 & 276 & 186 & 107 & 73 & 41 & 4\\
    dancetrack0029 & 3474 & 1088 & 385 & 86 & 8 & 0 & 0\\
    dancetrack0032 & 1015 & 190 & 177 & 132 & 75 & 10 & 2\\
    dancetrack0033 & 2819 & 530 & 339 & 209 & 170 & 29 & 3\\
    dancetrack0037 & 3025 & 392 & 360 & 184 & 134 & 17 & 2\\
    dancetrack0039 & 2115 & 195 & 103 & 42 & 4 & 1 & 0\\
    dancetrack0044 & 6379 & 817 & 459 & 230 & 95 & 15 & 1\\
    dancetrack0045 & 7271 & 1319 & 736 & 290 & 84 & 11 & 1\\
    dancetrack0049 & 3937 & 786 & 462 & 227 & 116 & 32 & 2\\
    dancetrack0051 & 6377 & 1247 & 816 & 328 & 69 & 11 & 0\\
    dancetrack0052 & 1293 & 166 & 100 & 95 & 64 & 15 & 1\\
    dancetrack0053 & 1147 & 159 & 92 & 53 & 22 & 1 & 0\\
    dancetrack0055 & 1219 & 176 & 97 & 56 & 35 & 6 & 0\\
    dancetrack0057 & 655 & 83 & 43 & 15 & 7 & 1 & 0\\
    dancetrack0061 & 1392 & 102 & 61 & 41 & 25 & 5 & 0\\
    dancetrack0062 & 1027 & 193 & 91 & 42 & 26 & 9 & 0\\
    dancetrack0066 & 1328 & 187 & 136 & 72 & 28 & 3 & 1\\
    dancetrack0068 & 2479 & 364 & 253 & 160 & 55 & 11 & 0\\
    dancetrack0069 & 2574 & 247 & 97 & 26 & 9 & 0 & 0\\
    dancetrack0072 & 1849 & 250 & 147 & 78 & 28 & 4 & 0\\
    dancetrack0074 & 2566 & 307 & 115 & 31 & 3 & 0 & 0\\
    dancetrack0075 & 1965 & 196 & 131 & 61 & 79 & 20 & 0\\
    dancetrack0080 & 3793 & 220 & 62 & 3 & 2 & 0 & 0\\
    dancetrack0082 & 6668 & 509 & 119 & 14 & 7 & 1 & 0\\
    dancetrack0083 & 9673 & 528 & 105 & 13 & 1 & 0 & 0\\
    dancetrack0086 & 8182 & 1112 & 531 & 184 & 51 & 0 & 0\\
    dancetrack0096 & 10401 & 1644 & 880 & 179 & 15 & 1 & 0\\
    dancetrack0098 & 4371 & 767 & 406 & 132 & 48 & 4 & 0\\
    dancetrack0099 & 5676 & 664 & 291 & 106 & 32 & 16 & 0\\
    \midrule
    Overall & 164091 & 22717 & 12360 & 5603 & 2303 & 465 & 31\\
    Average per frame & 3.9260 & 0.5435 & 0.2957 & 0.1341 & 0.0551 & 0.0111 & 0.0007\\
    \bottomrule
    \end{tabular}}
    \label{tabs3}
\end{table}

\section{Comparison with generative models}
Generative model~\cite{dhariwal2021diffusion,rombach2022high,song2020denoising,ho2020denoising} gets a lot of  attention because of the appearance of diffusion model, and the excellent performance in applications like AIGC makes the study~\cite{saharia2022photorealistic,ruiz2022dreambooth} of this technology more and more popular. Like our method, the diffusion models also uses the concept of "noise and denoise". These methods gradually add Gaussian noises to the original image until the image becomes a totally noisy image, then a network is used to gradually denoise the image until the image becomes the original image.

There are several differences between our method and the generative model like diffusion model:

\begin{itemize}
    \item \textbf{Purposes of noising.} Our purpose is to simulate the occurrence of occlusion by adding noises, while the diffusion model just adds noise to make the image gradually become pure Gaussian noise.
    \item \textbf{Format of noising.} Our noises are bounding boxes-level, while the noises of the diffusion model are pixel-level.
    \item \textbf{Procedures of denoise.} We use a decoder architecture directly regress the original bounding box or map the noising bounding boxes to the category of "no-object", while diffusion gradually restores the original image using the network repeatedly.
\end{itemize}

\section{comparison with methods in other fields which adopt ``noising-and-denoising'' strategy}

Compared to the generative model, there are also many discriminative models that follow the concept of ``noise and denoise". In this section, we compare the difference between our model and methods using denoising paradigm in other fields (e.g., DN-DETR~\cite{li2022dn} and DINO~\cite{zhang2022dino} in object detection, which is our baseline).

Though our method follows the negative noise and noise mask module proposed in those methods, we innovatively propose the cascaded mask and conditional noises according to the severe occlusions.  Overall, the differences between our method and others are as follows:
\begin{itemize}
    \item The generation of denoising queries in other methods is completely random, which lacks consideration for the surrounding environment, including objects of the same and different categories. In our method, we consider the complex environment around the tracking object, and according to whether there are other objects in the vicinity, propose innovative conditional noises, and simulate the ID switches occurred in the crowded scenes.
    \item We also use a Mask to coordinate different types of queries in decoder, but we mainly focus on track queries rather than denosing queries, because unlike other methods in other fields, the relationship among the track queries and the relationship between track queries and other queries is the key of MOT. Besides, we propose a Cascaded Mask Module to dynamically adapt our design of mask, so in different stages of decoder, the track queries can interact with each other and focus on themselves, respectively.
    \item Our method is the first to use the ``noise and denoise" paradigm in the multiple object tracking.
\end{itemize}

\section{Detailed experimental setup}
In Table \ref{tabs1}, we provide our hyperparameters in our experiments. Learning rate is initialized as $2\times10^{-4}$ and decays to $2\times10^{-5}$ after 10 epochs. We pretrain with 80 epochs in CrowdHuman and then train 40, 50 and 20 epochs in MOT17~\cite{milan2016mot16}, MOT20~\cite{dendorfer2020mot20} and DanceTrack~\cite{sun2022dancetrack}, respectively. The backbone is ResNet50 and we use 4 levels of feature map as mentioned in the main text. We adopt 6 layers of encoder layers and 6 layers of decoder layers while the hidden dimension is set to 288 and the number of head is set to 8. Because MOT20 has more objects than other datasets, we set the number of detection queries to 300 in MOT20 while 150 in others. We follow Deformable DETR and set the number of sampled points to 4 in both encoder and decoder. The total number of positive denoising queries is limited to 200 and we noise the calss ID up to 20. For the loss function, we set all the weight to 1 and among track loss, we set the weight of focal loss, L1 loss and IOU loss to 1.0, 5.0 and 2.0, respectively. The positive noises range factor is set to 0.2 and negative noises factor is set to 0.4, while in positive noises, if there exists condition positive noises, the probability is 0.6.
\begin{table}[]
    \centering
    \caption{Hyperparameter settings in our experiments.}
    \scalebox{1.0}{
    \begin{tabular}{lccc}
    \toprule    Parameter & MOT17 & MOT20 & DanceTrack \\
    \midrule
    $Learning\; rate\; (Transformer)$ & $2\times10^{-4}$ & $2\times10^{-4}$ & $2\times10^{-4}$\\
    $Learning\; rate\; (Backbone)$ & $2\times10^{-5}$ & $2\times10^{-5}$ & $2\times10^{-5}$\\
    $Epoch$ & 40 & 50 & 20 \\
    $Learning \;rate\; decay\; epoch$ & 10 & 10 & 10 \\
    $Learning\; rate\; decay\; rate$ & 0.1 & 0.1 & 0.1\\
    $Backbone$ & ResNet50 & ResNet50 & ResNet50\\
    $Number\; of\; feature\; levels$ & 4 & 4 & 4 \\
    $Number\; of\; encoder\; layers$ & 6 & 6 & 6\\
    $Number\; of\; decoder\; layers$ & 6 & 6 & 6\\
    $Hidden\_dim$ & 288 & 288 & 288 \\
    $Dropout\; rate$ & 0.1 & 0.1 & 0.1 \\
    $Number\; of\; heads\; in\; Attention$ & 8 & 8 & 8 \\
    $Number\; of\; detection\; queries$ & 150 & 300 & 150 \\
    $Number\; of\; sampled\; points\; in\; encoder$ & 4 & 4 & 4 \\
    $Number\; of\; sampled\; points\; in\; decoder$ & 4 & 4 & 4 \\
    $\lambda_\text{focal}$ & 1.0 & 1.0 & 1.0 \\
    $\lambda_\text{L1}$ & 5.0 & 5.0 & 5.0 \\
    $\lambda_\text{iou}$ & 2.0 & 2.0 & 2.0 \\
    $\lambda_\text{track}$ & 1.0 & 1.0 & 1.0 \\
    $\lambda_\text{aux}$ & 1.0 & 1.0 & 1.0 \\
    $\lambda_\text{inter}$ & 1.0 & 1.0 & 1.0 \\
    $\lambda_\text{r}$ & 0.2 & 0.2 & 0.2 \\
    $\lambda_\text{n}$ & 0.4 & 0.4 & 0.4 \\
    $\lambda_\text{c}$ & 0.4 & 0.6 & 0.6 \\
    $Category\; noise\; range$ & 20 & 20 & 20 \\
    $Number\; of\; Denoising\; queries$ & 200 & 200 & 200 \\
    \bottomrule
    \end{tabular}}

    \label{tabs1}
\end{table}

\section{Visualization}
\begin{figure}
  \centering
  \includegraphics[width=\linewidth]{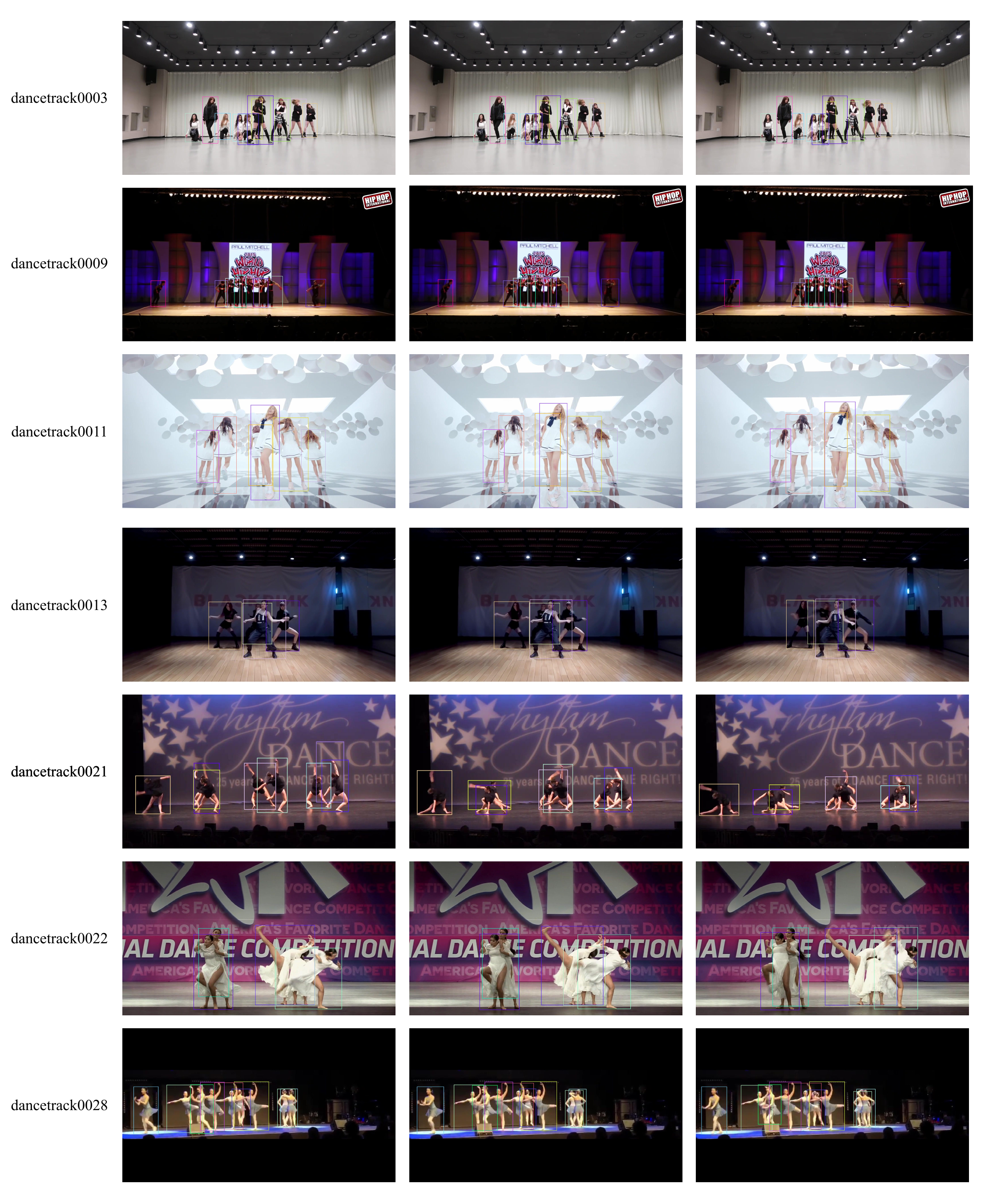}
  \caption{Visualization of sequences in DanceTrack test set.}
  \label{fig4}
\end{figure}

\begin{figure}
  \centering
  \includegraphics[width=\linewidth]{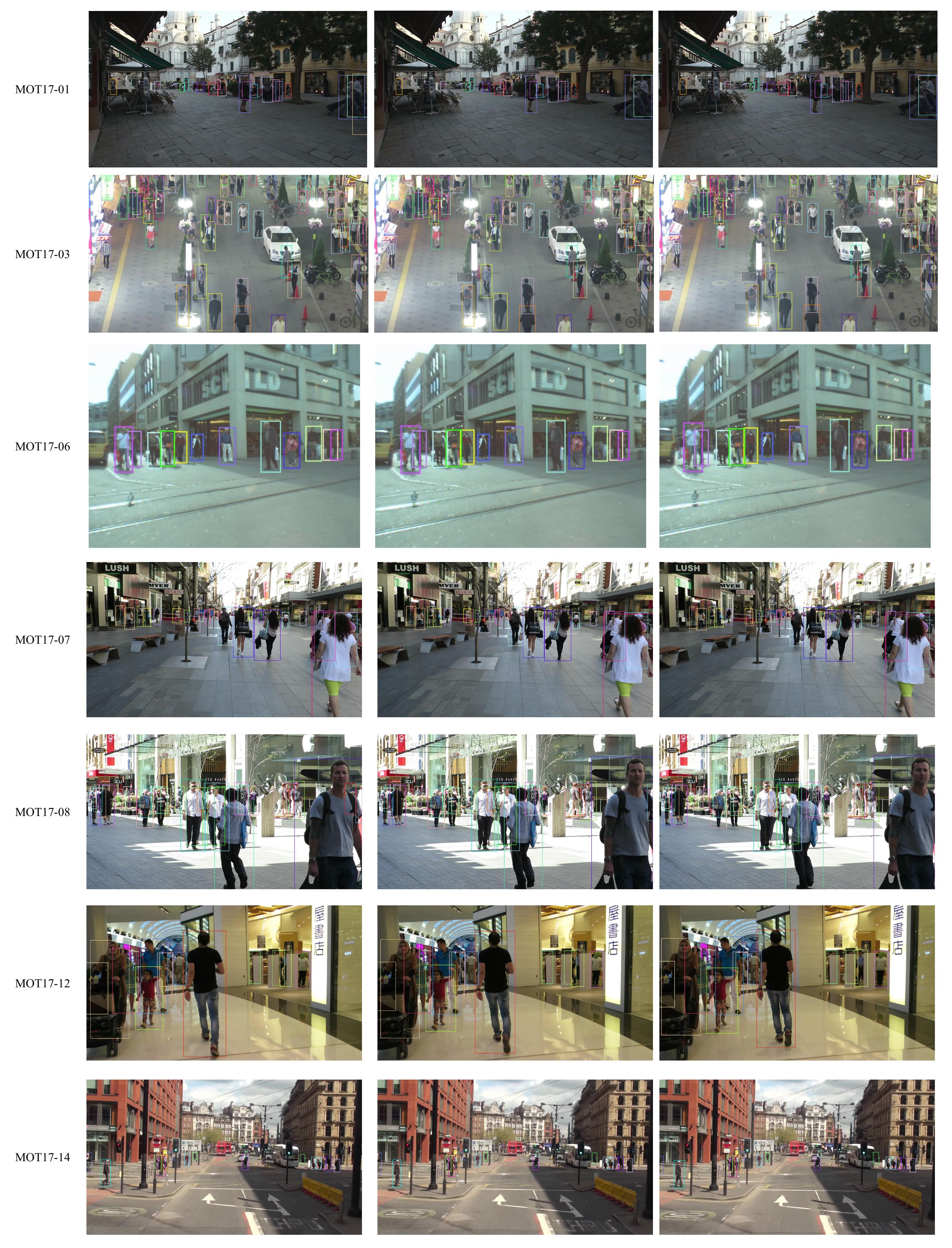}
  \caption{Visualization of sequences in MOT17 test set.}
  \label{fig2}
\end{figure}

\begin{figure}
  \centering
  \includegraphics[width=\linewidth]{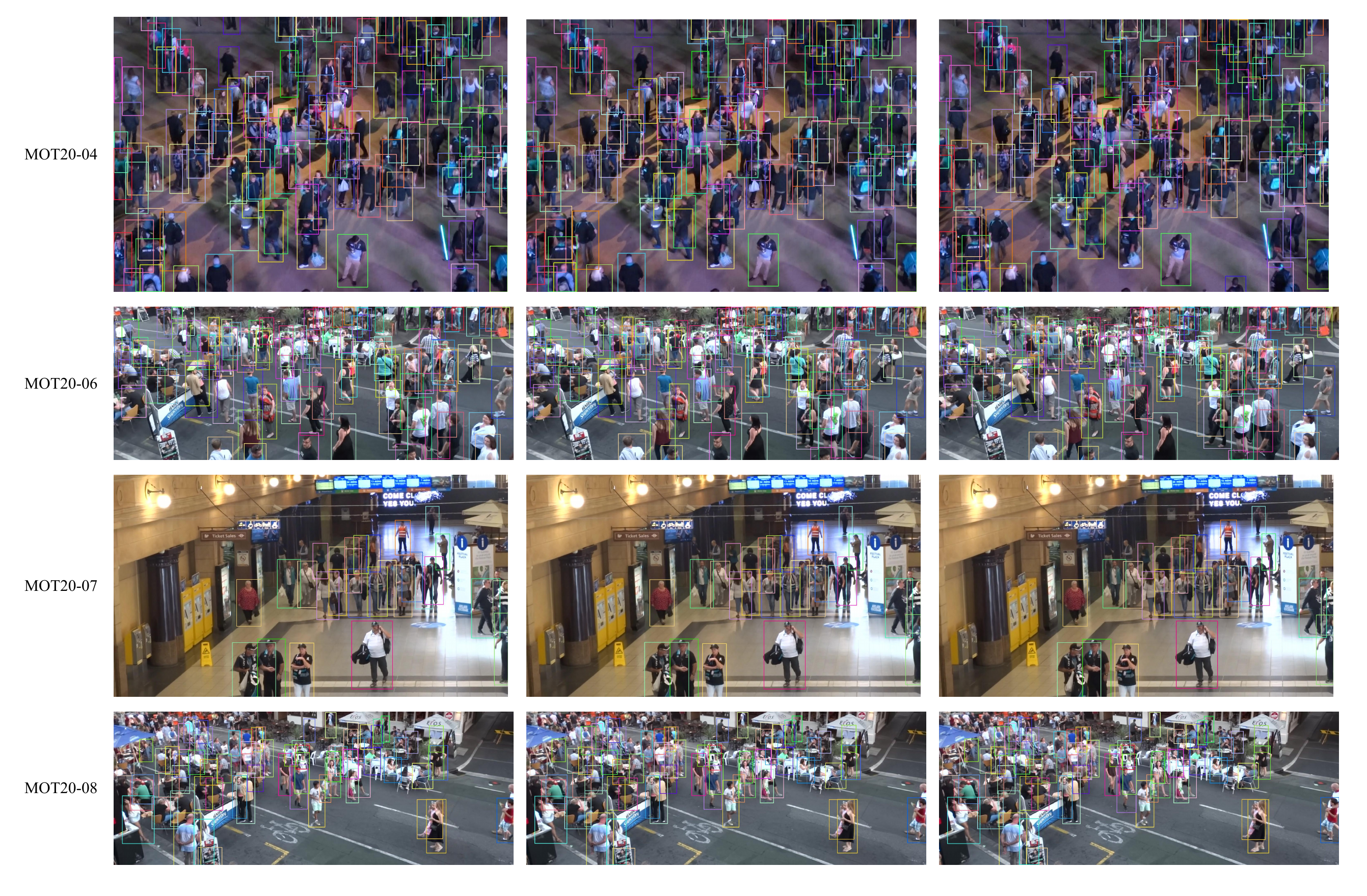}
  \caption{Visualization of sequences in MOT20 test set.}
  \label{fig3}
\end{figure}

In this section, we provide more\cite{zhang2022humandiffusion} visualization results. For MOT17 and MOT20 dataset, we randomly display three consecutive frames for all sequences in test set in Figure \ref{fig2} and Figure \ref{fig3}, respectively. For DanceTrack, there are 35 sequences in test set, so we select the first part of sequences and randomly display three consecutive frames for those sequences in Figure \ref{fig4}.

\end{sloppypar}
\end{document}